\documentclass[a4paper,twoside]{article}

\usepackage{graphicx}
\usepackage{float}
\usepackage{multirow}
\usepackage{amsmath,amssymb} % define this before the line numbering.
\usepackage{hyperref}
\usepackage{subcaption}
\usepackage{color,soul}
\usepackage{multirow}
\usepackage{rotating}

\usepackage{epsfig}
\usepackage{calc}
\usepackage{amssymb}
\usepackage{amstext}
\usepackage{amsmath}
\usepackage{amsthm}
\usepackage{multicol}
\usepackage{pslatex}
\usepackage{apalike}
\usepackage{SCITEPRESS}     % Please add other packages that you may need BEFORE the SCITEPRESS.sty package.

\begin{document}

\title{Optical Flow augmented Semantic Segmentation networks \\ for Automated Driving}

\author{\authorname{Hazem Rashed\sup{1}, Senthil Yogamani\sup{2}, Ahmad El-Sallab\sup{1}, Pavel K\v{r}\'{i}\v{z}ek\sup{3} and Mohamed El-Helw\sup{4}}
\affiliation{\sup{1}CDV AI Research, Cairo, Egypt}
\affiliation{\sup{2}Valeo Vision Systems, Ireland}
\affiliation{\sup{3}Valeo R\&D DVS, Prague, Czech Republic }
\affiliation{\sup{4}Nile University, Cairo, Egypt}
\email{\{hazem.rashed, senthil.yogamani, ahmad.el-sallab@valeo.com\}@valeo.com, melhelw@nu.edu.eg}
}


\abstract{Motion is a dominant cue in automated driving systems. Optical flow is typically computed to detect moving objects and to estimate depth using triangulation. In this paper, our motivation is to leverage the existing dense optical flow to improve the performance of semantic segmentation.   To provide a systematic study, we construct four different architectures which use RGB only, flow only, RGBF concatenated and two-stream RGB + flow. We evaluate these networks on two automotive datasets namely Virtual KITTI and Cityscapes using the state-of-the-art flow estimator FlowNet v2. We also make use of the ground truth optical flow in Virtual KITTI to serve as an ideal estimator and a standard Farneback optical flow algorithm to study the effect of noise.  Using the flow ground truth in Virtual KITTI, two-stream architecture achieves the best results with an improvement of 4\% IoU.  As expected, there is a large improvement for moving objects like trucks, vans and cars with 38\%, 28\% and 6\% increase in IoU. FlowNet produces an improvement of 2.4\% in average IoU with larger improvement in the moving objects corresponding to 26\%, 11\% and 5\% in trucks, vans and cars. In Cityscapes, flow augmentation provided an improvement for moving objects like motorcycle and train with an increase of 17\% and 7\% in IoU.}

\onecolumn \maketitle \normalsize \vfill

\section{Introduction}
Semantic image segmentation has witnessed tremendous progress recently with deep learning. It provides dense pixel-wise labeling of the image which leads to scene understanding. Automated driving is one of the main application areas where it is commonly used \cite{horgan2015vision}. The level of maturity in this domain has rapidly grown recently and the computational power of embedded systems have increased as well to enable commercial deployment. Currently, the main challenge is the cost of constructing large datasets as pixel-wise annotation is very labor intensive. It is also difficult to perform corner case mining as it is a unified model to detect all the objects in the scene. Thus there is a lot of research to reduce the sample complexity of segmentation networks by incorporating domain knowledge and other cues where-ever possible. In this work, we explore the usage of motion cues via dense optical flow to improve the accuracy.

% \begin{figure}[!t]
% \centering
% \includegraphics[width=0.45\textwidth]{Images/segnet}
% \caption{Semantic Segmentation of a typical automotive scene}
% \label{fig:segnet}
% \end{figure}

Majority of semantic segmentation algorithms mainly rely on appearance cues based on a single image and do not exploit motion cues from two consecutive images. 
%Figure \ref{fig:segnet} illustrates semantic segmentation of a typical automotive scene computed from RGB. 
In this paper, we address the usage of dense optical flow as a motion cue in semantic segmentation networks. In particular for automotive scenes, the scene is typically split into static infrastructure and set of independently moving objects. Motion cues which needs two consecutive frames could particularly improve the segmentation of moving objects.

% \begin{figure*}[!t]
% \centering
% \includegraphics[width=\textwidth]{Images/DeepFlow}
% \caption{Dense optical flow estimation on a typical automotive scene}
% \label{fig:flow}
% \end{figure*}

The contributions of this work include: 

\begin{itemize}
    \item Construction of four CNN architectures to systematically study the effect of optical flow augmentation to semantic segmentation.
    \item Experimentation on two automotive datasets namely Virtual KITTI and Cityscapes.
    \item Ablation study on using different flow estimators and different flow representations.
\end{itemize}

The rest of the paper is organized as follows: Section \ref{sec:related} reviews the related work in segmentation, computation of flow and role of flow in semantic segmentation. Section \ref{sec:method} details the construction of four architectures to systematically study the effect of augmenting flow to semantic segmentation networks. Section \ref{sec:exps} discusses the experimental results in Virtual KITTI and Cityscapes. Finally, section \ref{sec:conc} provides concluding remarks.

\section{Related Work}
\label{sec:related}
\subsection{Semantic Segmentation}

A detailed survey of semantic segmentation for automated driving is presented in \cite{siam2017deep}. We briefly summarize the relevant parts focused on CNN based methods which are split into mainly three subcategories. The first \cite{farabet2013learning} used patch-wise training to yield the final classification. In\cite{farabet2013learning} an image is fed into a Laplacian pyramid, each scale is forwarded through a 3-stage network to extract hierarchical features and patch-wise classification is used. The output is post processed with a graph based classical segmentation method. In \cite{grangier2009deep} a deep network was used for the final pixel-wise classification to alleviate any post processing needed. However, it still utilized patch-wise training. 

The second subcategory \cite{long2015fully}\cite{noh2015learning}\cite{badrinarayanan2015segnet} was focused on end-to-end learning of pixel-wise classification.  It started with the work in \cite{long2015fully} that developed fully convolutional networks (FCN). The network learned heatmaps that was then upsampled with-in the network using deconvolution to get dense predictions. Unlike patch-wise training methods this method uses the full image to infer dense predictions.  In \cite{noh2015learning} a deeper deconvolution network was developed, in which stacked deconvolution and unpooling layers are used. In Segnet \cite{badrinarayanan2015segnet} a similar approach was used where an encoder-decoder architecture was deployed. The decoder network upsampled the feature maps by keeping the maxpooling indices from the corresponding encoder layer. 

% Finally, the work in \cite{yu2015multi}\cite{farabet2013learning}\cite{noh2015learning}\cite{chen2015attention}\cite{Qi_2016_CVPR}\cite{ronneberger2015u} focused on multiscale semantic segmentation. Initially in \cite{farabet2013learning} the scale issue was addressed by introducing multiple rescaled versions of the image to the network. However with the emergence of end-to-end learning, the skip-net architecture in \cite{long2015fully} was used to merge heatmaps from different resolutions. Since these architectures rely on downsampling the image, loss of resolution can hurt the final prediction. The work in \cite{ronneberger2015u} proposed a u-shaped architecture network where feature maps from different initial layers are upsampled and concatenated for the next layers. Another work in \cite{yu2015multi} introduced dilated convolutions, which expanded the receptive field without losing resolution based on the dilation factor. Thus it provided a better solution for handling multiple scales. Finally the recent work in \cite{chen2015attention} provided a better way for handling scale. It uses attention models that provides a mean to focus on the most relevant features with-in the image. This attention model is able to learn a weighting map that weighs feature maps pixel-by-pixel from different scales.

\subsection{Optical Flow in Automated Driving Systems}

Flow estimation is very critical for automated driving and it has been a standard module for more than ten years. Due to computational restriction, sparse optical flow was used and it is replaced by Dense Optical Flow (DOF) in recent times. As flow is already computed, it can be leveraged for semantic segmentation. Motion estimation is a challenging problem because of the continuous camera motion along with the motion of independent objects. Moving objects are the most critical in terms of avoiding fatalities and enabling smooth maneuvering and braking of the car. Motion cues can also enable generic object detection as it is not possible to train for all possible object categories beforehand. Classical approaches in motion detection were focused on geometry based approaches \cite{torr1998geometric}\cite{papazoglou2013fast}\cite{ochs2014segmentation}\cite{menze2015object}\cite{scott2017motion}. However, pure geometry based approaches have many limitations, motion parallax issue is one such example. A recent trend  \cite{tokmakov2016learning}\cite{jain2017fusionseg}\cite{drayer2016object}\cite{vijayanarasimhan2017sfm}\cite{fragkiadaki2015learning} for learning motion in videos has emerged. CNN based optical flow has produced state of the art results. FlowNet \cite{ilg2016flownet} was a simple two stream structure which was trained on synthetic data. 
% Figure \ref{fig:flow} illustrates the visualization of FlowNet computed for an automotive scene. In this image, the optical flow corresponding to moving pedestrian is distinct and thus can benefit semantic segmentation of pedestrians.

There has been many attempts to combine appearance and motion cues for various tasks. Jain et. al. presented a method for appearance and motion fusion in \cite{jain2017fusionseg} for generic foreground object segmentation. Laura et. al. \cite{sevilla2016optical} leverages semantic segmentation for customizing motion model for various objects. This has been commonly used in scene flow models. Junhwa et. al \cite{hur2016joint} uses optical flow to provide temporally consistent semantic segmentation via post processing. MODNet \cite{siam2017modnet} fuses optical flow and RGB images for moving object detection. FuseNet \cite{hazirbas2016fusenet} is the closest to the work in this paper where they augmented semantic segmentation networks with depth. They show that concatenation of RGBD slightly reduces mean IoU and two-stream network with cross links show an improvement of 3.65 \% IoU in SUN RBG-D dataset. Motion is a complementary cue of color and its role is relatively less explored for semantic segmentation. Our motivation is to construct simple baseline architectures which can be built on top of current CNN based semantic segmentation. More sophisticated flow augmentation architectures were proposed in \cite{nilsson2016semantic} and \cite{gadde2017semantic}, however they are computationally more intensive.

\section{Semantic Segmentation Models}
\label{sec:method}
\begin{figure}[ht!]
\centering
\begin{subfigure}{0.5\textwidth}
    \includegraphics[scale=0.495]{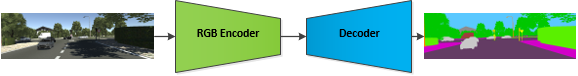}
    \caption{\textcolor{black}{Input RGB\newline}}
\end{subfigure}%
\qquad
\begin{subfigure}{0.5\textwidth}
    \includegraphics[scale=0.495]{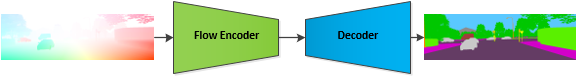}
    \caption{\textcolor{black}{Input Flow\newline}}    
\end{subfigure}%
\qquad
\begin{subfigure}{0.5\textwidth}
    \includegraphics[scale=0.495]{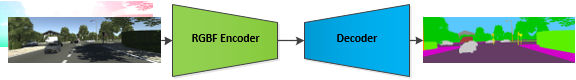}
    \caption{\textcolor{black}{Input RGBF\newline}}    
\end{subfigure}%
\qquad
\begin{subfigure}{0.5\textwidth}
    \includegraphics[scale=0.495]{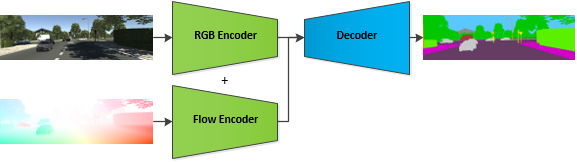}
    \caption{\textcolor{black}{Two Stream RGB+F\newline}}    
\end{subfigure}%

\caption{ Four types of architectures constructed and tested in the paper. (a) and (b) are baselines using RGB and Flow only. (c) and (d) are flow augmented semantic segmentation architectures.}
\label{fig:architectures}
\end{figure}

In this section, we discuss the details of the four different semantic segmentation networks used in this paper. We construct two flow augmented architectures namely RGBF network (Figure \ref{fig:architectures} (c)) which does concatenation and two-stream RGB+F network (Figure \ref{fig:architectures} (d)). RGB only and flow only architectures serve as baselines for comparative analysis. 

\begin{table}[ht!]
\tiny
\centering
\caption{Quantitative evaluation on Virtual KITTI data for our four segmentation networks. }
\begin{tabular}{|l|l|l|l|l|}
\hline
Network Type & IoU & Precision & Recall & F-Score \\ \hline
RGB & 66.47	& 78.23 & 75.6	& 73.7\\ \hline
F & 42	& 63.92	& 55.28	& 55.76\\ \hline
RGBF (concat) & 65.33 &	82.37 & 73.77 & 75.85 \\ \hline
RGB + F (2-stream add) & \textbf{70.52} & \textbf{83} & \textbf{76.4} & \textbf{78.9} \\ \hline
\end{tabular}
\label{table:VKitti_Mean_Metrics}
\end{table}

\begin{table*}[ht!]
\tiny

%%%%%%%%%%%%%%%%%%%%%%%%%%%%%%%%%%%%%%%%%%%%%%%%%%%%%
%Optical flow representation tables

\bigskip
\centering
\caption{Semantic Segmentation Results on Virtual Kitti (Mean IoU) for different DOF (GT) representations}
\label{table:VKitti_Eval_DOF_Representations}
\begin{tabular}{|c|c|c|c|c|c|c|c|c|c|c|c|c|}
\hline
Type  & Mean & Truck	& Car &	Van  &	Road &	Sky &	Vegetation &	Building	 & Guardrail &	TrafficSign &	TrafficLight &	Pole \\ \hline

RGBF (GT-Color Wheel)	& 59.88 & 	41.7 & 	84.44 & 	40.74 & 	93.76 & 	93.6 & 	66.3 & 	49.43 & 	52.18 & 	61.21 & 49.61 & 	21.52 
 \\ \hline

RGBF (GT-Mag \& Dir) &	58.82 & 	45.12 & 	82.3 & 	30.04 & 	90.25 & 	\textbf{94.1} & 	60.97 & 	56.48 & 	51.48 & 	58.74 & 	49.7 & 	26.01 \\ \hline 

RGBF (GT-Mag only) & 65.32 & 	70.73 & 	80.16 & 	48.33 & 	93.59 & 	93.3 & 	70.79 & 	62.04 & 	67.86 & 	55.13 & 	55.48 & 	31.92 \\ \hline

RGB+F (GT-3 layers Mag only)	& 67.88 & 	35.7 & 	91.02 & 	24.78 & 	\textbf{96.47} & 	94.06 & 	\textbf{88.72} & 	\textbf{74.4} & 	\textbf{84.5} & 	\textbf{69.48} & 	\textbf{68.95} & 	\textbf{34.28} 
\\ \hline

RGB+F (GT-Color Wheel) & \textbf{70.52} &	\textbf{71.79} &	\textbf{91.4} &	\textbf{56.8}	& 96.19	& 93.5 & 83.4 &	66.53 & 82.6 &	64.69 &	64.65	& 26.6 \\ \hline

\end{tabular}

\bigskip
\centering
\caption{Semantic Segmentation Results (Mean IoU) on CityScapes for different DOF (Farneback) representations}
\label{table:CityScapes_Eval_DOF_Representations}
\begin{tabular}{|c|c|c|c|c|c|c|c|c|c|c|c|c|c|}
\hline
Type & 	Mean	&  Bicycle & 	Person & 	Rider & 	Motorcycle	&  Bus & 	Car	& Train & 	Building & 	Road & 	 {Truck} & 	Sky & TrafficSign\\ \hline

RGBF (Mag only) &	47.8 & 	52.63 & 	55.82 & 	31.08 & 	22.38 & 	39.34 & 	82.75 & 	22.8 & 	80.43 & 	92.24 & 	 {20.7} & 	81.87 & 	44.08
\\ \hline

RGBF (Mag \& Dir) &	54.6 & 	57.28 & 	58.63 & 	33.56 & 	18.49 & 	56.44 & 	87.6 & 	41.15 & 	84.41 & 	95.4 & 	 {31.8} & 	87.86 & 	44.26
\\ \hline

RGBF (Color Wheel) &	57.2 & 	61.47 & 	62.18 & 	35.13 & 	22.68 & 	54.87 & 	87.45 & 	36.69 & 	86.28 & 	95.94 & 	 {40.2} & 	90.07 & 	51.64
\\ \hline

RGB+F (3 layers Mag only)	& 62.1 & 	\textbf{65.15} & 	65.44 & 	32.59 & 	33.19 & 	63.07 & 	89.48 & 	43.6 & 	\textbf{87.88} & 	96.17 & 	 {\textbf{57.2}} & 	\textbf{91.48} & 	55.76 \\ \hline

RGB+F (Color Wheel) & \textbf{62.56}	& 63.65	& \textbf{66.3} &	\textbf{39.65}	& \textbf{47.22}	& \textbf{66.24}	& \textbf{89.63}	& \textbf{51.02}	& 87.13	& \textbf{96.4} &	 {36.11}	& 90.64 & \textbf{60.68} \\ \hline

\end{tabular}
\bigskip

%%%%%%%%%%%%%%%%%%%%%%%%%%%%%%%%%%%%%%%%%%%%%%%%%%

\centering
\caption{Semantic Segmentation Results (Mean IoU) on Virtual KITTI dataset}
\label{table:VKitti_Eval_SepClasses}
\begin{tabular}{|c|c|c|c|c|c|c|c|c|c|c|c|c|}
\hline
Type  & Mean & Truck	& Car &	Van  &	Road &	Sky &	Vegetation &	Building	 & Guardrail &	TrafficSign &	TrafficLight &	Pole \\ \hline
RGB & 66.47 &	33.66	& 85.69	 & 29.04 &	95.91 &	\textbf{93.91} & 80.92	& 68.15 & 	81.82	& \textbf{66.01}	& \textbf{65.07} &	\textbf{40.91} \\ \hline
F (GT) & 42 &	36.2 &	55.2	& 20.7 &	62.6 &	93.9 &	19.54 &	34 &	15.23 &	51.5 &	33.2 &	29.3 \\ \hline
RGBF (GT) & 65.328 &	70.74	& 80.2 &	48.34	& 93.6	& 93.3	 & 70.79 &	62.05 &	67.86 &	55.14	& 55.48	& 31.9 \\ \hline
RGB+F (GT) & \textbf{70.52} &	\textbf{71.79} &	\textbf{91.4} &	\textbf{56.8}	& \textbf{96.19}	& 93.5 & 83.4 &	66.53 & \textbf{82.6} &	64.69 &	64.65	& 26.6 \\ \hline
F (Flownet) & 28.6 & 24.6 &	47.8 &	14.3 & 57.9	& 68 & 13.4 & 4.9 & 0.8 & 31.8 & 18.5 & 6.6 \\ \hline
RGB+F (Flownet) & 68.84 &	60.05 &	90.87 &	40.54	& 96.05	& 91.73 & \textbf{84.54} &	\textbf{68.52} & 82.43 &	65.2 &	63.54	& 26.54 \\ \hline
\end{tabular}

\bigskip

\centering
\caption{Semantic Segmentation Results (Mean IoU) on Cityscapes dataset}
\label{table:Cityscapes}
\begin{tabular}{|c|c|c|c|c|c|c|c|c|c|c|c|c|c|}
\hline
Type & 	Mean	&  Bicycle & 	Person & 	Rider & 	Motorcycle	&  Bus & 	Car	& Train & 	Building & 	Road & 	 {Truck} & 	Sky & TrafficSign \\ \hline

RGB & 62.47	& 63.52	&  \textbf{67.93}	& 40.49	& 29.96	& 62.13	& 89.16	& 44.19	& \textbf{87.86}	& 96.22	&  {48.54}	& 89.79 & 59.88 \\ \hline

F (Farneback) &  34.7 &	34.48 &	37.9	& 12.7 &	7.39 &	31.4 &	74.3 &	11.35 &	72.77 &	91.2 &	 {19.42} &	79.6 & 11.4 \\ \hline
RGBF (Farneback) & 47.8 & 52.6	& 55.8	& 31.1	& 22.4	& 39.34	& 82.75	& 22.8	& 80.43	& 92.24	&  {20.7}	& 81.87	& 44.08 \\ \hline

RGB+F (Farneback) & \textbf{62.56}	& 63.65	& 66.3 &	39.65	& \textbf{47.22}	& \textbf{66.24}	& 89.63	& \textbf{51.02}	& 87.13	& \textbf{96.4} &	 {36.1}	& 90.64 & \textbf{60.68} \\ \hline

F (Flownet) & 36.8	& 32.9	& 50.9 &	26.8	& 5.12	& 25.99	& 75.29	& 15.1	& 65.16	& 90.75 &	 {25.46}	& 50.16 & 29.14 \\ \hline

RGBF (Flownet) & 52.3	& 54.9	& 58.9 &	34.8	& 26.1	& 53.7	& 83.6	& 40.7	& 79.4	& 94 &	 28.1	& 79.4 & 45.5 \\ \hline

RGB+F (Flownet) & 62.43	& \textbf{64.2}	& 66.32 &	\textbf{40.9}	& 40.76	& 66.05	& \textbf{90.03}	& 41.3	& 87.3	& 95.8 &	 {\textbf{54.7}}	& \textbf{91.07} & 58.21 \\ \hline
\end{tabular}

\end{table*}

\subsection{One-stream networks}
An encoder-decoder architecture is used for performing semantic segmentation. Our network is based on FCN8s \cite{long2015fully} architecture, after which the remaining fully connected layers of the VGG16 are transformed to a fully convolutional network. The first 15 convolutional layers are used for feature extraction. 
The segmentation decoder follows the FCN architecture. 1x1
convolutional layer is used followed by three transposed
convolution layers to perform up-sampling. Skip
connections are utilized to extract high resolution features from
the lower layers and then the extracted features are added to the partially upsampled results. The exact same one stream architecture is used for both RGB-only and flow-only experiments.
We use the cross-entropy loss function as shown below.
% \begin{equation}
% L= - \frac{1}{|I|} \sum_{i \in I} \sum_{c \in C_{Dataset}} p_i(c)\log{q_i(c)}
% \end{equation}
where q denotes predictions and p denotes ground-truth. $C_{Dataset}$ is the set of classes for the used dataset.

\subsection{RGBF network}
The input to this network is a 3D volume containing the original RGB image concatenated with the flow map. Several optical flow representations have been studied experimentally to find the best performing input, namely Color wheel representation in 3 channels, magnitude and direction in 2 channels, and magnitude only in 1 channel. It was found that an input of 4 channels containing RGB image concatenated with optical flow magnitude performs the best (Refer to Table \ref{table:VKitti_Eval_DOF_Representations} and \ref{table:CityScapes_Eval_DOF_Representations}). The flow map is normalized from 0 to 255 to have the same value range as the RGB. The first layer of the VGG is adapted to use the input of four channels and the corresponding weights are initialized randomly. The rest of the network utilizes the VGG pre-trained weights. In case of Virtual KITTI, we make use of flow ground truth as one of the inputs. This is done to simulate a near-perfect flow estimator which can be achieved by state-of-the-art CNN based algorithms and eliminate the variability due to estimation errors. In case of cityscapes, we make use of OpenCV Farneback function where the magnitude of the flow vectors are computed as \(\sqrt{u^2 + v^2}\) where u and v are the horizontal and vertical flow vectors output from the function. The magnitude is then normalized to 0-255. In both the datasets, we also make use of the state-of-the-art flow estimator FlowNet v2 \cite{ilg2017flownet}.

\subsection{Two stream (RGB+F) network}
Inspired from \cite{simonyan2014two}\cite{jain2017fusionseg}\cite{siam2017modnet}, we construct a two stream network which utilizes two parallel VGG16 encoders to extract appearance and flow features separately. The feature maps from both streams are fused together using summation junction producing encoded features of same dimensions. Then the same decoder described in the one-stream network is used to perform upsampling. Following the same approach as the one-stream network, skip connections are used to benefit from the high resolution feature maps. This network produces the best performance among the four as discussed in the experiments section. However, the two stream network is computationally more complex with more parameters relative to the RGBF network. The main difference is the fusion at the encoder stage rather than early fusion in RGBF network. RGB+F network also has the advantage of being able to re-use pre-trained image encoders as it is decoupled from flow.

\section{Experiments}
\label{sec:exps}
In this section, we present the datasets used, experimental setup and results. 

\subsection{Datasets}

We had the following criteria for choosing the semantic segmentation datasets for our experiments. Firstly, the dataset has to be for automotive road scenes as our application is focused on automated driving. Secondly, we needed a mechanism to have the previous image for computational of optical flow. We chose two datasets namely 
Virtual KITTI \cite{Gaidon:Virtual:CVPR2016} and Cityscapes \cite{cordts2016cityscapes} datasets which satisfy the above criteria. Cityscapes is a popular automotive semantic segmentation dataset developed by Daimler. It comprises of 5000 images having full semantic segmentation annotation and 20,000 images with coarse annotations. We only use the fine annotations in our experiment and evaluated performance on the validation set that comprises of 500 images. We also chose Virtual KITTI dataset for obtaining ground truth for flow which can be used as a proxy for a perfect flow estimator. It is a synthetic dataset that consists of 21,260 frames generated using Unity game engine in urban environment under different weather conditions. Virtual KITTI provides many annotations of which two are utilized in our approach namely dense flow and semantic segmentation.  We split the dataset into 80\% training data and 20\% testing data. The semantic information is given for 14 classes including static and moving objects.

\subsection{Experimental Setup}
For all experiments, Adam optimizer is used with a learning rate of $1e^{-5}$. L2 regularization is used in the loss function with a factor of value of $5e^{-4}$ to avoid over-fitting the data. The encoder is initialized with VGG pre-trained weights on ImageNet. Dropout with probability 0.5 is utilized for 1x1 convolutional layers. Input image resolution of 375x1242 is used for Virtual KITTI and 1024x2048 is used for Cityscapes which is downsized to 512x1024 during training.
Intersection over Union (IoU) is used as the performance metric for each class separately and an average IoU is computed over all the classes. In addition to IoU, precision, recall and F-score were calculated for Virtual KITTI dataset. 

\subsection{Experimental Results}

We provide several tables of qualitative evaluation on the two datasets and discuss the impact of various classes due to flow augmentation. We also provide video links on outputs of the four architectures on test sequences in both datasets. In case of Virtual KITTI, we use the synthetic flow annotation as input. This is used to simulate a perfect flow estimator so that it can act as a best case baseline. Then it is compared with FlowNet v2 algorithm.  In case of Cityscapes, we use a commonly used dense optical flow estimation algorithm namely Farneback's algorithm in OpenCV 3.1 with default parameters. We intentionally use this algorithm to understand the effects of relatively noisy flow estimations and compare it with FlowNet v2.

Table \ref{table:VKitti_Mean_Metrics} shows the evaluation of our four architectures on Virtual KITTI dataset with different metrics. Flow augmentation consistently provides improvement in all four metrics. There is an improvement of 4\% in IoU, 4.3\% in Precision, 4.36\% in Recall and 5.89\% in F-score. Tables \ref{table:VKitti_Eval_DOF_Representations} and  \ref{table:CityScapes_Eval_DOF_Representations} shows a quantitative comparison of different DOF representations in RGBF and RGB+F architectures for both VKITTI and Cityscapes datasets. Results show that Color Wheel based representation in RGB+F architecture provides the best capability of capturing the motion cues and thus the best semantic segmentation accuracy. For all the following experiments, we make use of Color Wheel representation format for flow. \\

\begin{figure*}[ht!]
\centering
\begin{subfigure}{.48\textwidth}
    \includegraphics[width=\textwidth]{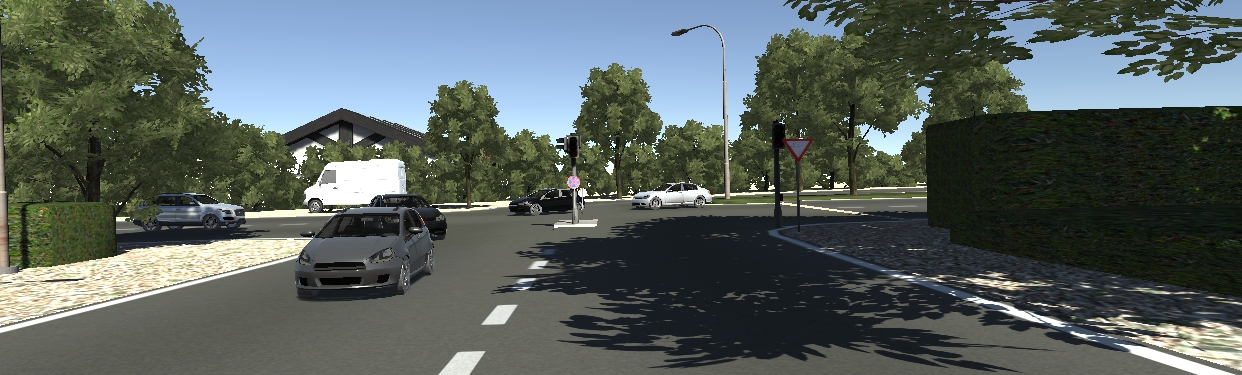}
    \caption{\textcolor{black}{Input Image}\newline}
\end{subfigure}%
\quad
\begin{subfigure}{.48\textwidth}
    \includegraphics[width=\textwidth]{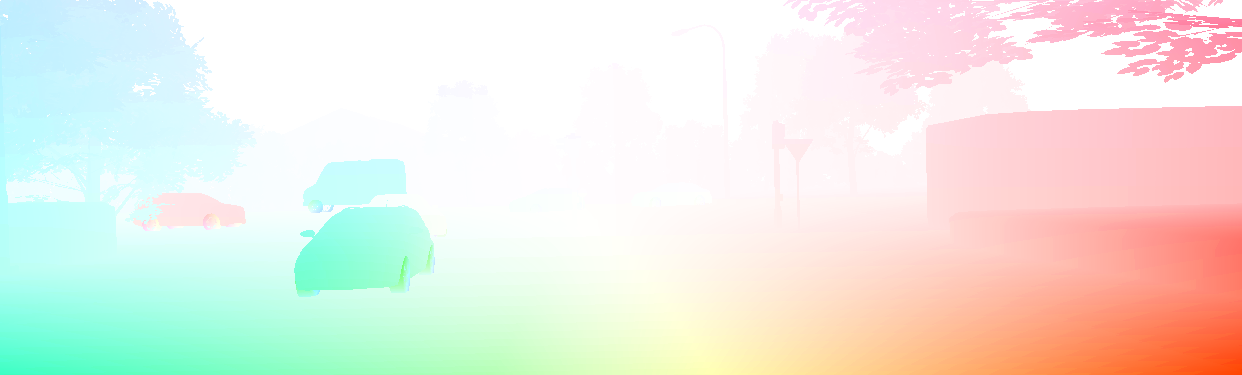}
    \caption{\textcolor{black}{Input DOF (Ground Truth)}\newline}
\end{subfigure}%
\quad
\begin{subfigure}{.48\textwidth}
    \includegraphics[width=\textwidth]{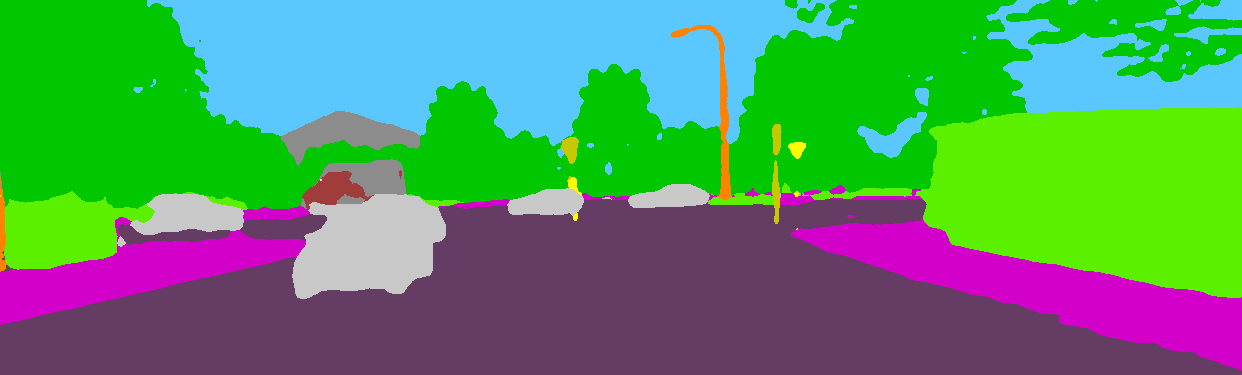}
    \caption{\textcolor{black}{RGB only} \newline}
\end{subfigure}%
\quad
\begin{subfigure}{.48\textwidth}
    \includegraphics[width=\textwidth]{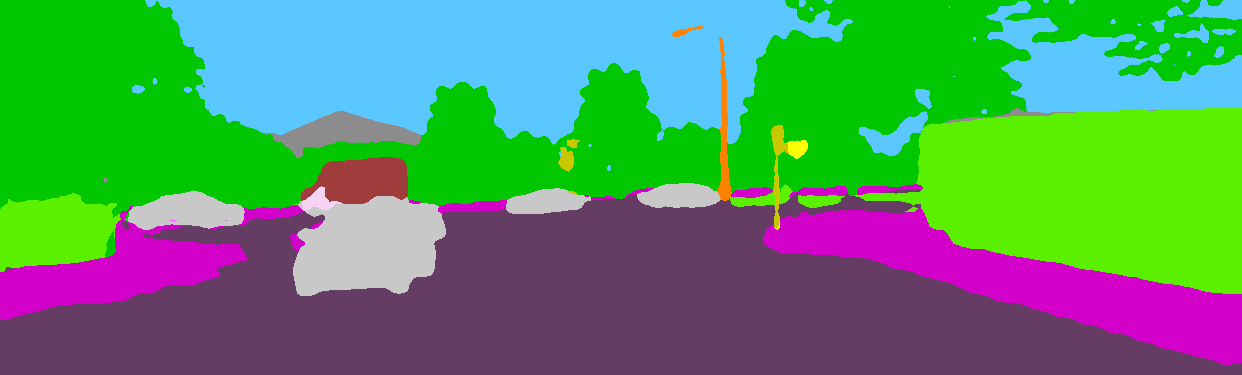}
    \caption{\textcolor{black}{Flow only (Ground Truth)}\newline}
\end{subfigure}

\begin{subfigure}{.48\textwidth}
    \includegraphics[width=\textwidth]{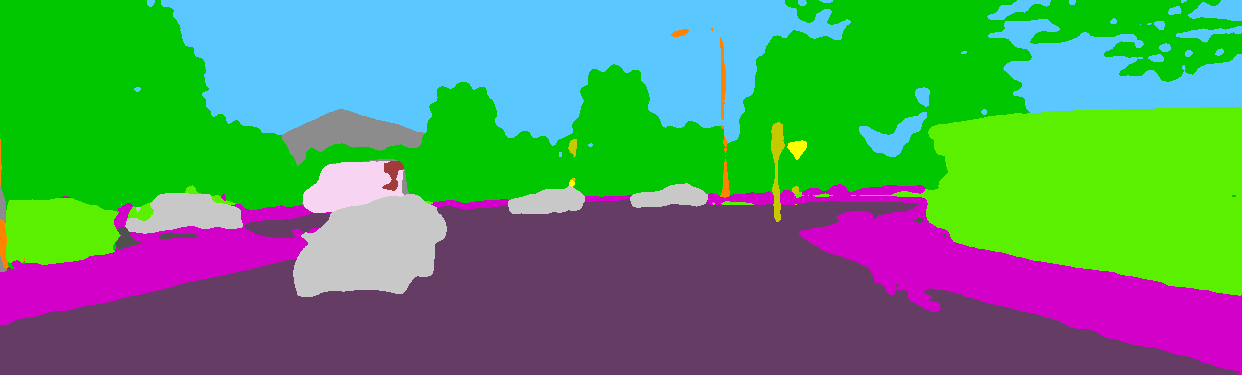}
    \caption{\textcolor{black}{RGBF (Ground Truth)}\newline}
\end{subfigure}%
\quad
\begin{subfigure}{.48\textwidth}
    \includegraphics[width=\textwidth]{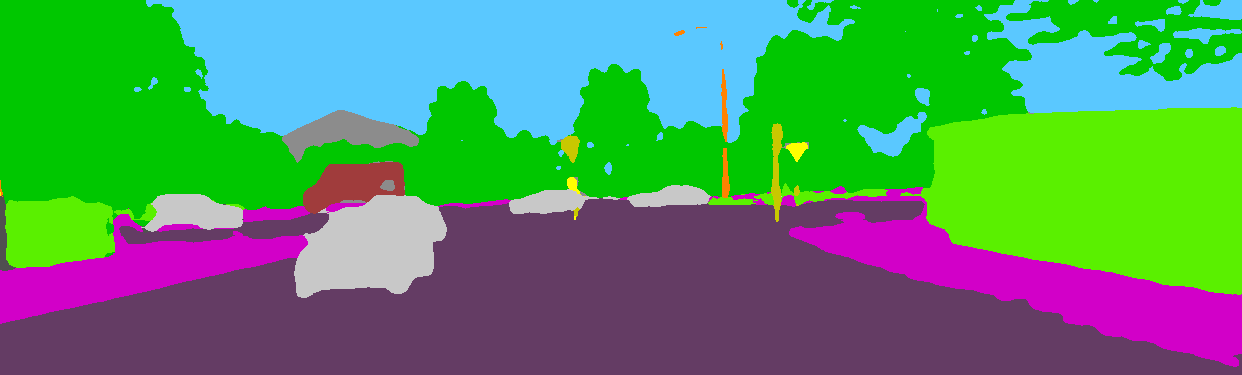}
    \caption{\textcolor{black}{RGB + F (Ground Truth)}\newline}
\end{subfigure}
\begin{subfigure}{.48\textwidth}
    \includegraphics[width=\textwidth]{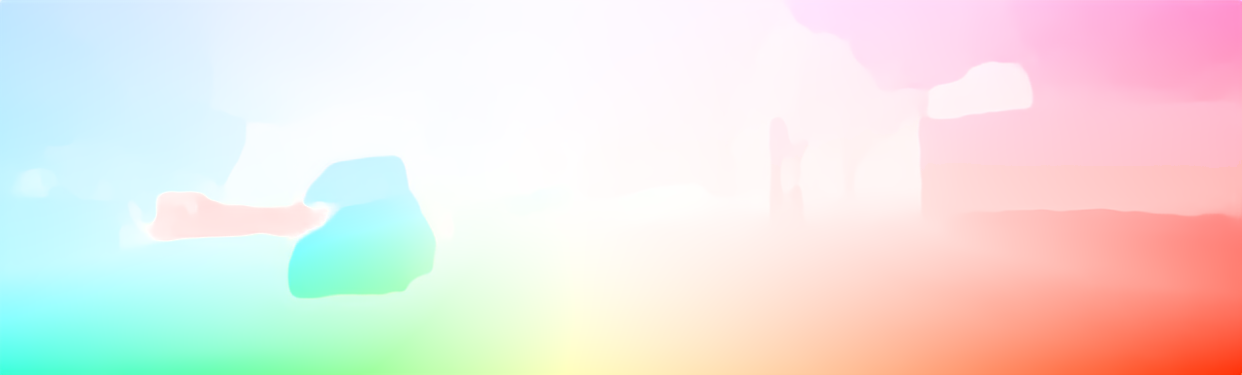}
    \caption{\textcolor{black}{Input DOF ( FlowNet)}\newline}
\end{subfigure}%
\quad
\begin{subfigure}{.48\textwidth}
    \includegraphics[width=\textwidth]{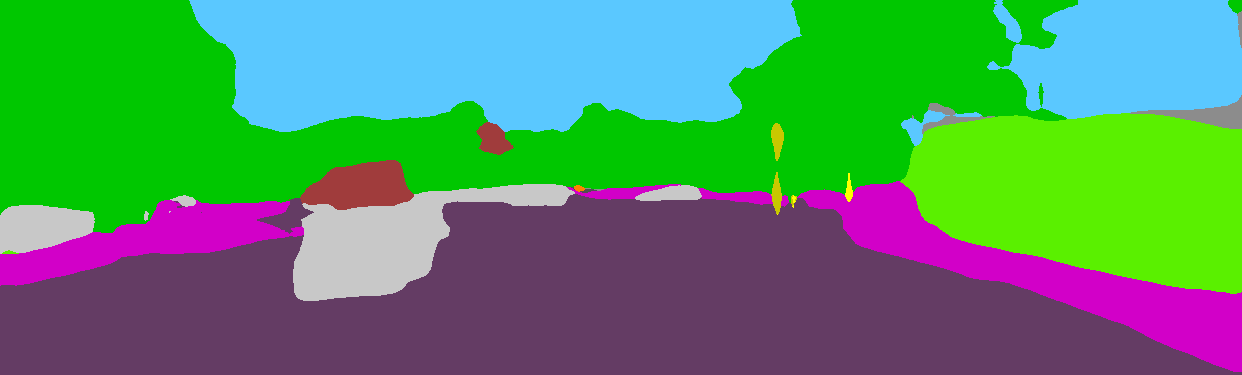}
    \caption{\textcolor{black}{Flow only ( FlowNet)}\newline}
\end{subfigure}%
\quad
\begin{subfigure}{.48\textwidth}
    \includegraphics[width=\textwidth]{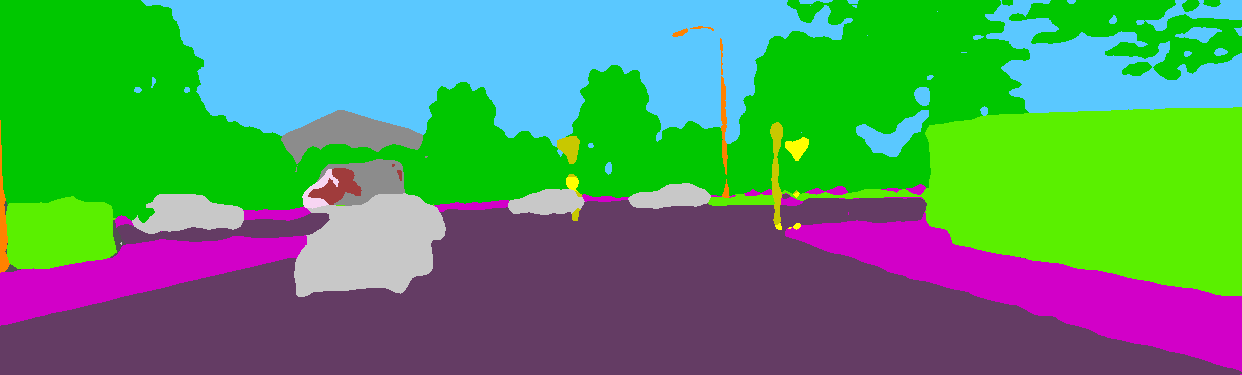}
    \caption{\textcolor{black}{RGB+F ( FlowNet)}}
\end{subfigure}%
\quad
\begin{subfigure}{.48\textwidth}
    \includegraphics[width=\textwidth]{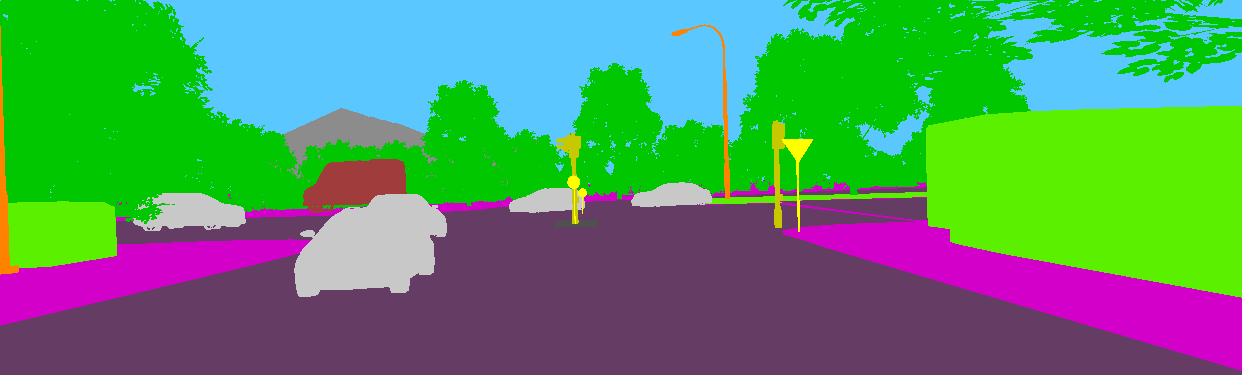}
    \caption{\textcolor{black}{Ground Truth} }
\end{subfigure}%
\quad

    \caption{Qualitative comparison of semantic segmentation outputs from four architectures  on Virtual KITTI dataset}
    \label{fig:comp_KITTI}
\end{figure*}

\begin{figure*}[htbp]
\centering
\begin{subfigure}{.48\textwidth}
    \includegraphics[width=\textwidth]{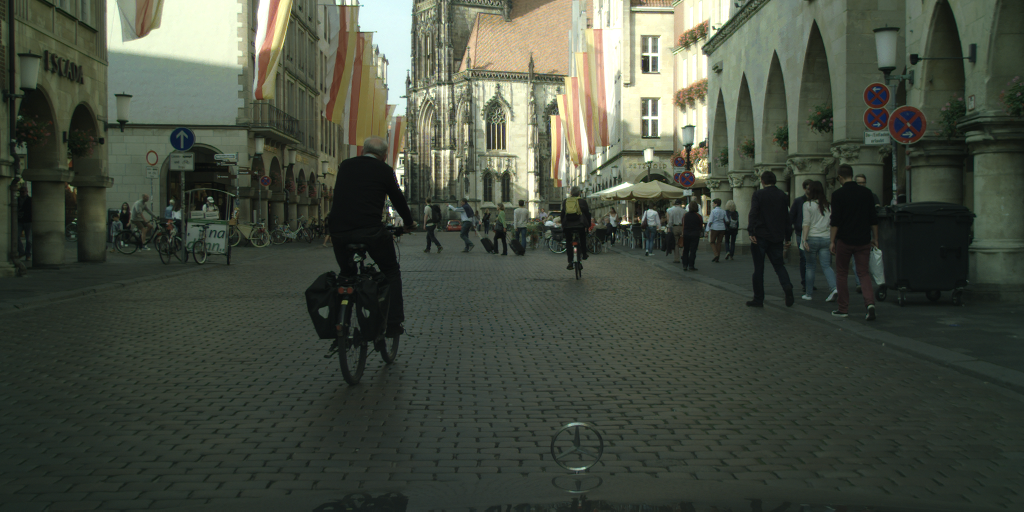}
    \caption{\textcolor{black}{Input Image}\newline}
\end{subfigure}%
% \quad
\begin{subfigure}{.48\textwidth}
    \includegraphics[width=\textwidth]{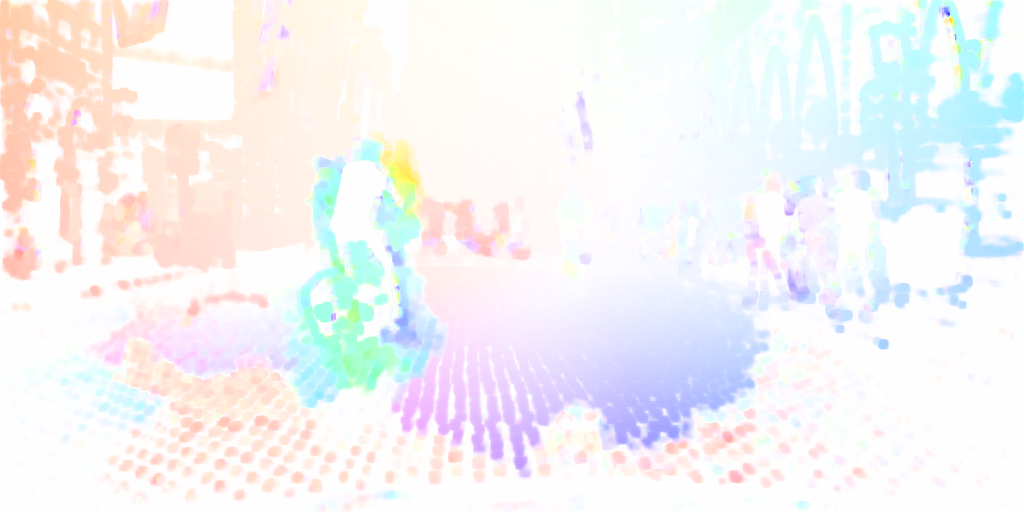}
    \caption{\textcolor{black}{Input DOF (Farneback)}\newline}
\end{subfigure}

\begin{subfigure}{.48\textwidth}
    \includegraphics[width=\textwidth]{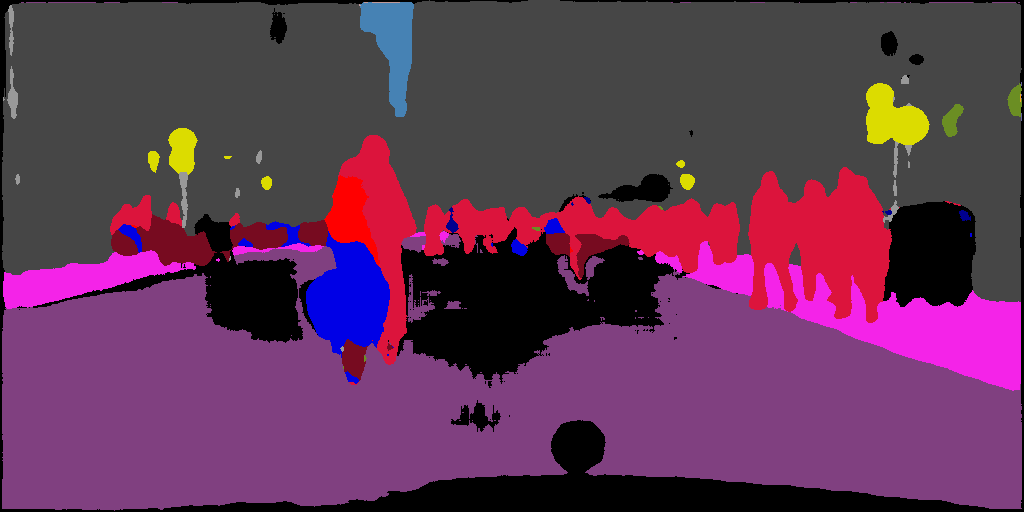}
    \caption{\textcolor{black}{RGB only output}\newline}
\end{subfigure}%
% \quad
\begin{subfigure}{.48\textwidth}
    \includegraphics[width=\textwidth]{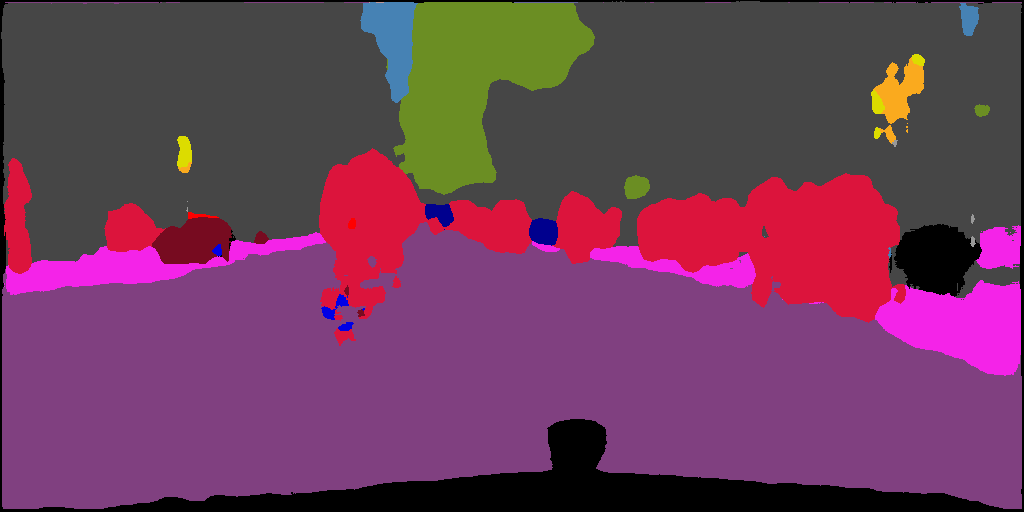}
    \caption{\textcolor{black}{Flow only (Farneback)}\newline}
\end{subfigure}

\begin{subfigure}{.48\textwidth}
    \includegraphics[width=\textwidth]{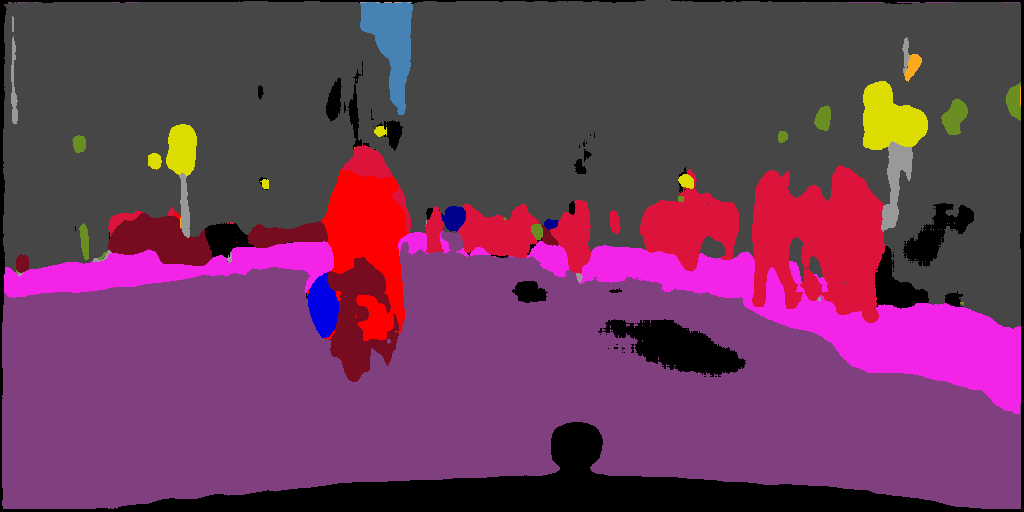}
    \caption{\textcolor{black}{RGBF output (Farneback)}\newline}
\end{subfigure}%
% \quad
\begin{subfigure}{.48\textwidth}
    \includegraphics[width=\textwidth]{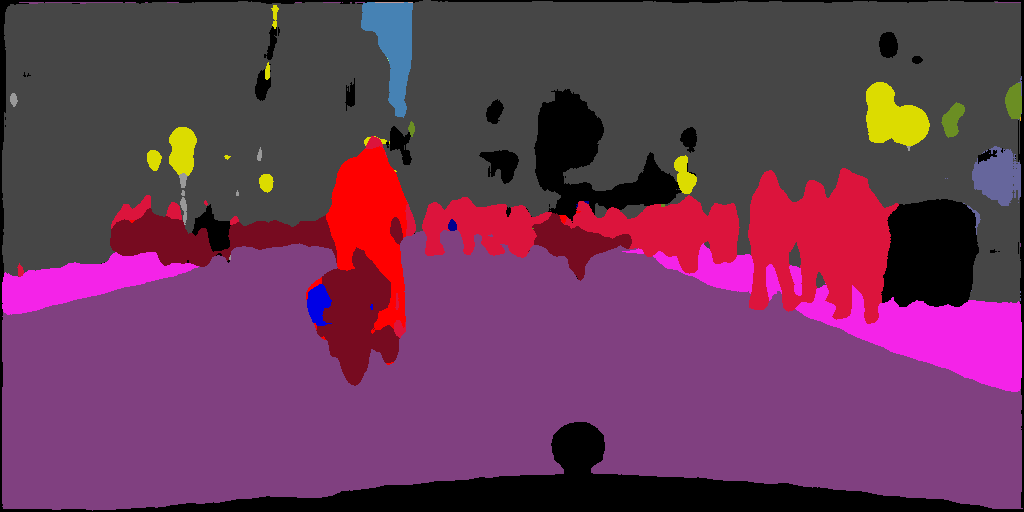}
    \caption{\textcolor{black}{RGB + F (Farneback)}\newline}
\end{subfigure}
\quad
\begin{subfigure}{.48\textwidth}
    \includegraphics[width=\textwidth]{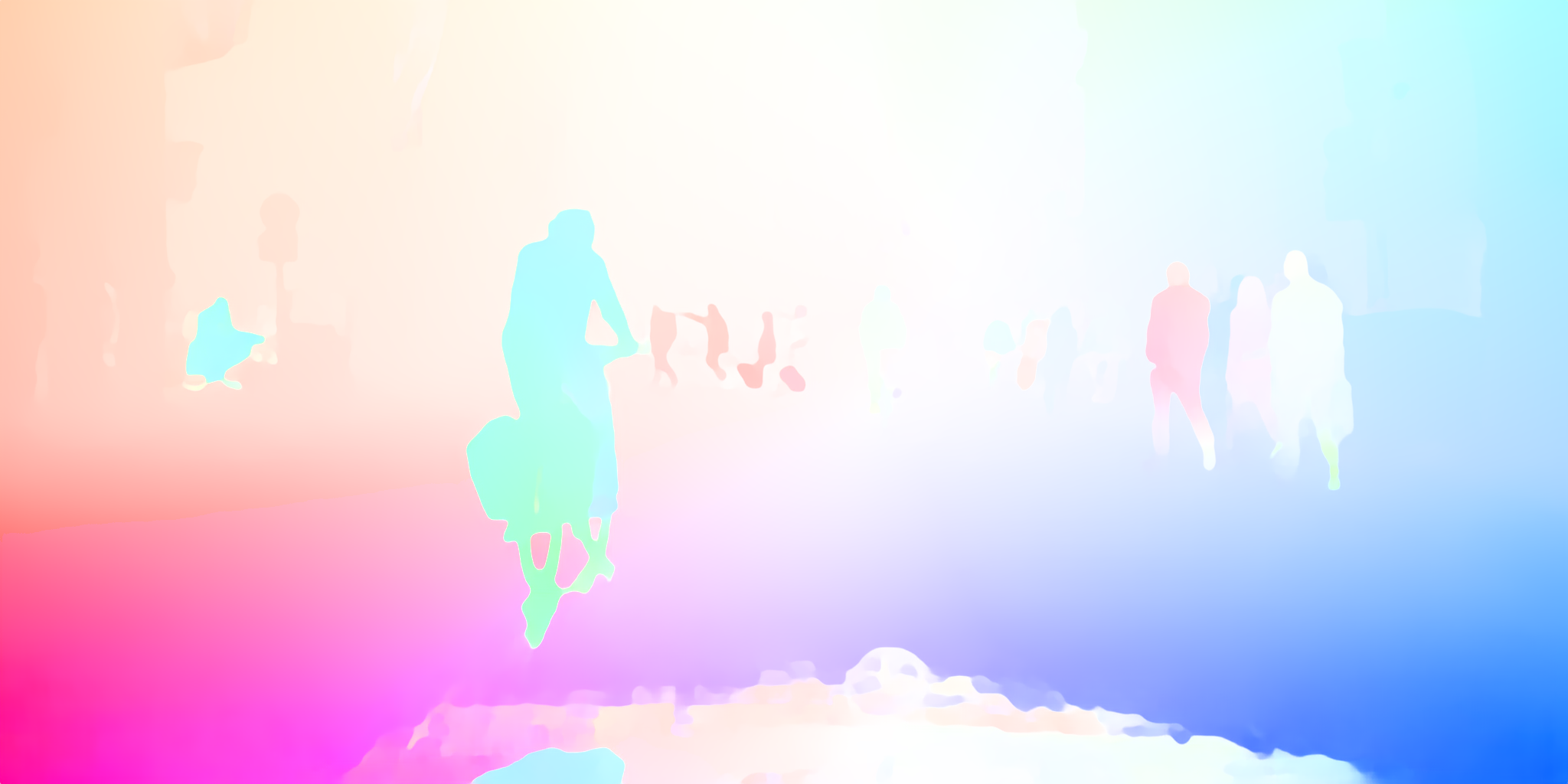}
    \caption{\textcolor{black}{Input DOF ( FlowNet)}\newline}
\end{subfigure}%
\begin{subfigure}{.48\textwidth}
    \includegraphics[width=\textwidth]{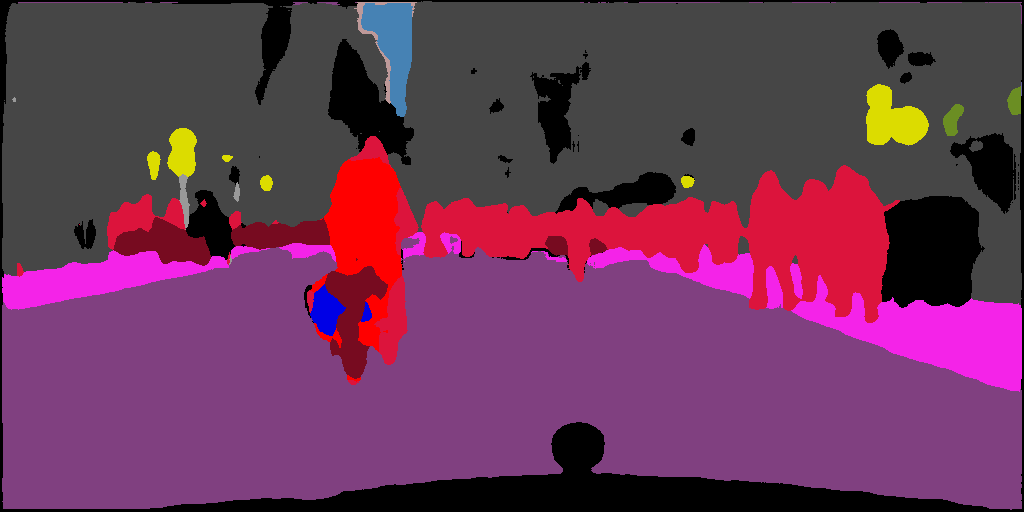}
    \caption{\textcolor{black}{RGB + F ( FlowNet)}\newline}
\end{subfigure}%
\quad
\begin{subfigure}{.48\textwidth}
    \includegraphics[width=\textwidth]{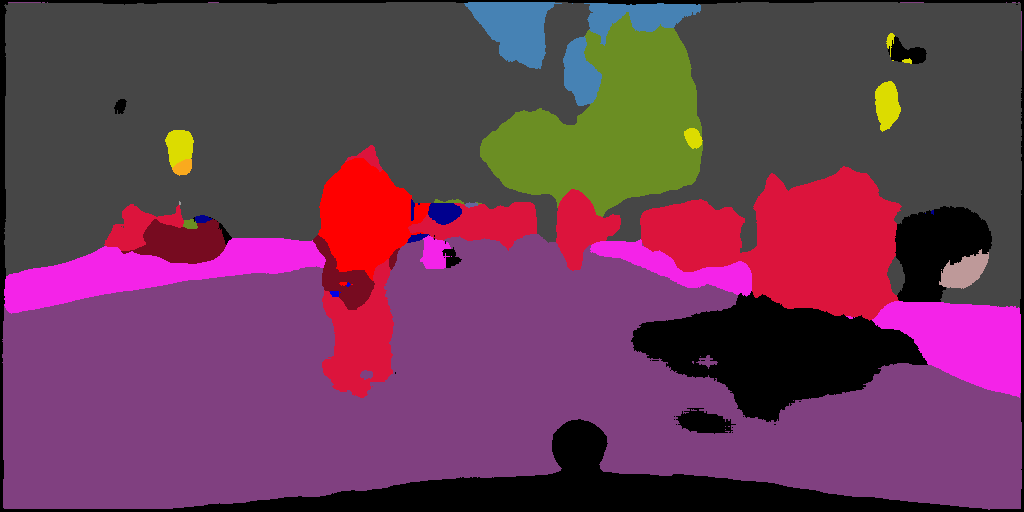}
    \caption{\textcolor{black}{DOF only (FlowNet)}\newline}
\end{subfigure}%
\begin{subfigure}{.48\textwidth}
    \includegraphics[width=\textwidth]{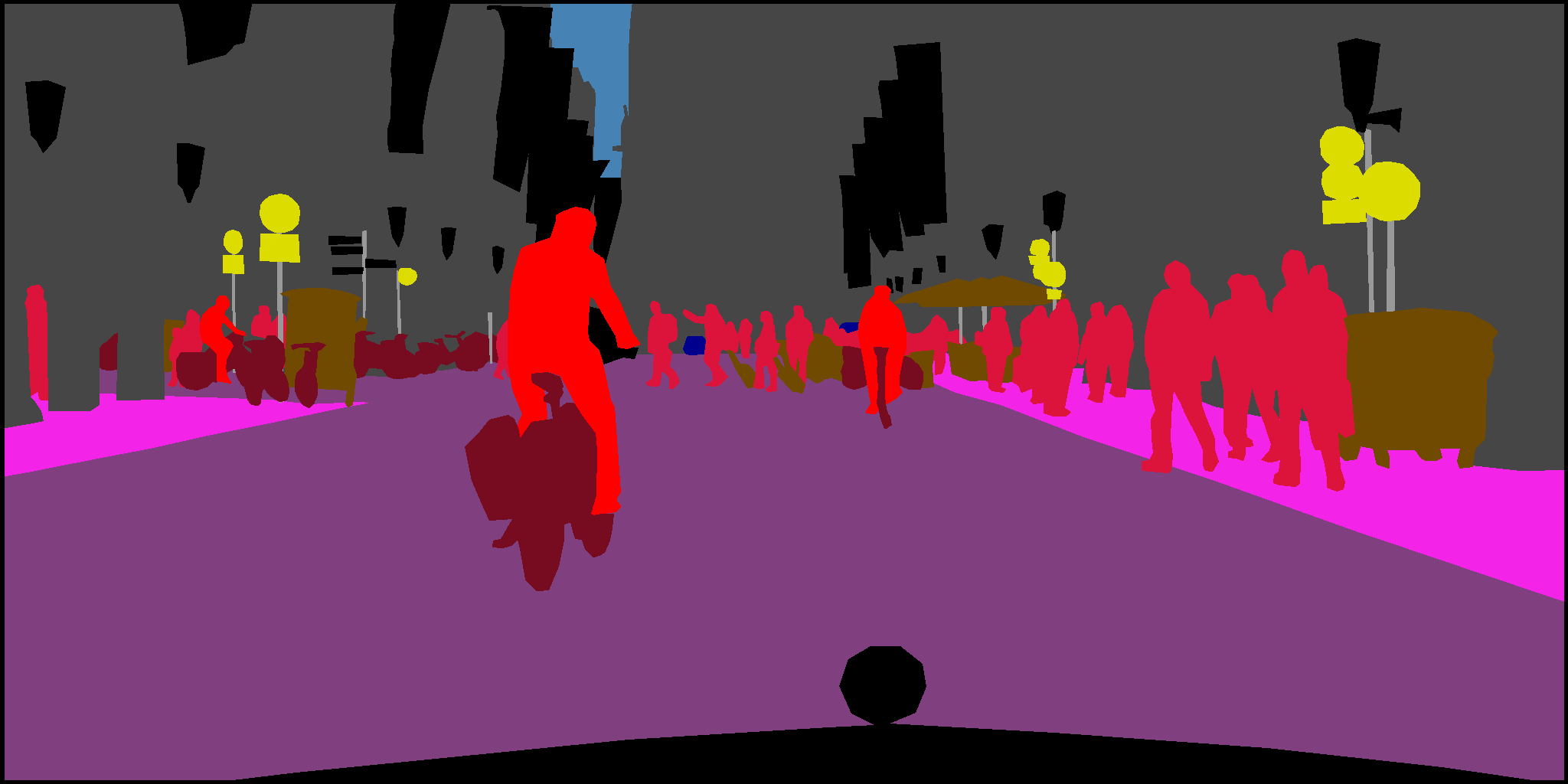}
    \caption{\textcolor{black}{Ground Truth}\newline}
\end{subfigure}
% \begin{subfigure}{.48\textwidth}
%     \includegraphics[width=\textwidth]{Images/munster_144_out_RGBF_flownet}
%     \caption{\textcolor{black}{RGBF (Flownet)}\newline}
% \end{subfigure}

    \caption{ Qualitative comparison of semantic segmentation outputs from four architectures  on  Cityscapes dataset }
    \label{fig:comp_Cityscapes}
\end{figure*}

Table \ref{table:VKitti_Eval_SepClasses} shows detailed evaluation for each class separately. Although the overall improvement is incremental, there is a large improvement for certain classes, for example, trucks, van, car show an improvement of 38.14\%, 27.8\% and 5.8\% respectively. Table \ref{table:Cityscapes} shows the evaluation of our proposed algorithm on Cityscapes dataset and shows a marginal improvement of 0.1\% in overall IoU. However there is a significant improvement in moving object classes like motorcycle and train by 17\% and 7\% even using noisy flow maps. It is important to note here that the average IoU is dominated by road and sky pixels as there is no pixel frequency based normalization and the obtained improvement in moving object classes is still significant. Figure \ref{fig:comp_KITTI} and Figure \ref{fig:comp_Cityscapes} illustrates qualitative results of the four architectures on Virtual KITTI and Cityscapes respectively. Figure \ref{fig:comp_KITTI} (f) shows better detection of the van which has a uniform texture and flow cue has helped to detect it better. Figure \ref{fig:comp_KITTI} (h) shows that flownet provides good segmentation for the moving van, however fusion in Figure \ref{fig:comp_KITTI} (i) still needs to be improved. Figure \ref{fig:comp_Cityscapes} (f) and (h) illustrate better detection of the bicycle and the cyclist after augmenting DOF. These examples visually verify the accuracy improvement obtained as shown in Table \ref{table:VKitti_Eval_SepClasses} and Table \ref{table:Cityscapes}. Unexpectedly, Table \ref{table:Cityscapes} shows that Farneback and FlowNet v2 provides similar results, however FlowNet shows better results for some classes like Truck. One of the design goals in this work is to have a computationally efficient model, our most complex architecture namely RGB+F runs at 4 fps on a TitanX Maxwell architecture GPU. Autonomous driving systems need real-time performance and better run-time can be obtained by using a more efficient encoder instead of the standard encoder we used in our experiments.

Semantic segmentation results in test sequences of both datasets for all the four architectures along with ground truth and original images are shared publicly in YouTube. The Virtual KITTI results are available in \footnote{https://youtu.be/4M1lS2-2w5U} and Cityscapes results are available in \footnote{https://youtu.be/VtFLpklatrQ}. We included flow only network to conduct an ablation study to understand what performance flow alone can produce. To our surprise, it produces good results especially for road, vegetation, vehicles and pedestrians. We noticed that there is degradation of accuracy relative to RGB baseline whenever there is noisy flow. It would be interesting to incorporate a graceful degradation via a loose coupling of flow in the network. In spite of flow only network showing good performance, the joint network only shows incremental improvement. It is likely that the simple networks we constructed do not effectively combine flow and RGB as they are completely different modalities with  different properties. It could also be possible that the RGB only network is implicitly learning geometric flow cues. In our future work, we plan to understand this better and construct better multi-modal architectures to effectively combine color and flow. 

% \begin{figure}[ht!]
% \begin{subfigure}{0.5\textwidth}
%     \includegraphics[scale= 0.45]{Images/sep-train}
% \end{subfigure}

% \begin{subfigure}{1.0\textwidth}
%     \includegraphics[scale= 1.0]{Images/joint-train}
% \end{subfigure}
% \end{figure}

% \begin{figure*}[ht!]
% \begin{subfigure}{.5\textwidth}
%     \includegraphics[scale= 0.55]{Images/1}
% \end{subfigure}%
% \begin{subfigure}{.5\textwidth}
%     \includegraphics[scale= 0.55]{Images/2}
% \end{subfigure}

% \begin{subfigure}{.5\textwidth}
%     \includegraphics[scale= 0.55]{Images/3}
% \end{subfigure}%
% \begin{subfigure}{.5\textwidth}
%     \includegraphics[scale= 0.55]{Images/4}
% \end{subfigure}

% \begin{subfigure}{.5\textwidth}
%     \includegraphics[scale= 0.55]{Images/5}
% \end{subfigure}%
% \begin{subfigure}{.5\textwidth}
%     \includegraphics[scale= 0.55]{Images/6}
% \end{subfigure}
%     \caption{Qualitative evaluation on KITTI MOD data for our proposed two-stream multi-task learning network MODNet. top row: Input Optical Flow, middle row output of 2 tasks: overlay motion mask (green), bottom row output of 3 tasks: overlay motion mask (yellow), road segmentation (green) and detected bounding boxes (blue).}
%     \label{fig:qual_seg}
% \end{figure*}

% \begin{figure*}[ht!]
% \centering
% \begin{subfigure}{.48\textwidth}
%     \includegraphics[width=\textwidth]{Images/0001_clone_00358_input_flownet}
%     \caption{\textcolor{black}{Input DOF-flownet}}
% \end{subfigure}%
% \quad
% \begin{subfigure}{.48\textwidth}
%     \includegraphics[width=\textwidth]{Images/0001_clone_00358_out_TwoStream_flownet}
%     \caption{\textcolor{black}{RGB+DOF (flownet)}}
% \end{subfigure}%
% \caption{Qualitative comparison of semantic segmentation output on Virtual KITTI dataset using flownet}
%     \label{fig:comp_KITTI}
% \end{figure*}

 \subsection{Future Work}
 
The scope of this work is to construct simple architectures to demonstrate the benefit of flow augmentation to standard CNN based semantic segmentation networks. The improvement in accuracy obtained and visual verification of test sequences shows that there is still a lot of scope for improvement. Flow and color are different modalities and an explicit synergistic model would probably produce better results compared to learning their relationships from data. We summarize the list of future work below: \\ (1) Understand the effect of different encoders of varying complexity like Resnet-10, Resnet-101, etc. \\
(2) Evaluation of state-of-the-art CNN based flow estimators and joint multi-task learning. \\
(3) Construction of better multi-modal fusion architectures. \\
(4) Auxiliary loss induced flow feature learning in semantic segmentation architecture. \\
(5) Incorporating sparsity invariant metrics to handle missing flow estimates.

\section{Conclusion}
\label{sec:conc}
In this paper, we explored the problem of leveraging flow in semantic segmentation networks. In applications like automated driving, flow is already computed in geometric vision pipeline and can be utilized. We constructed four variants of semantic segmentation networks which use RGB only, flow only, RGBF concatenated and two-stream RGB and flow. We evaluated these networks on two automotive datasets namely Virtual KITTI and Cityscapes. We provided class-wise accuracy scores and discussed them qualitatively. The simple flow augmentation architectures demonstrate a good improvement for moving object classes which are important for automated driving.  We hope that this study encourages further research in construction of better flow-aware networks to fully utilize its complementary nature.

%\vfill
\bibliographystyle{apalike}
{\small
\bibliography{IEEEfull}
}

%\section*{\uppercase{Appendix}}

%\noindent If any, the appendix should appear directly after the
%references without numbering, and not on a new page. To do so please use the following %command:
%\textit{$\backslash$section*\{APPENDIX\}}

\vfill
\end{document}